\documentclass{article}
\usepackage{spconf,amsmath,graphicx}
\usepackage{caption}
\usepackage{hyperref, setspace, lipsum}
\usepackage{multirow}
\usepackage{color, xcolor, colortbl}
\usepackage[sorting = none, backend = bibtex, style=numeric-comp]{biblatex}
\AtBeginBibliography{\small}
\usepackage{url}
\usepackage{subcaption}
\usepackage{float}
\usepackage{nccmath}
\usepackage{amssymb,amsthm,bbm}

\definecolor{darkred}{rgb}{0.7, 0, 0}
\definecolor{darkblue}{rgb}{0.0, 0.0, 0.7}
\definecolor{darkgreen}{rgb}{0, 0.4, 0}

\title{Wavelet-based Unsupervised Label-to-Image Translation}
\name{George Eskandar, Mohamed Abdelsamad, Karim Armanious, Shuai Zhang, Bin Yang}

\address{University of Stuttgart, Institute of Signal Processing and System Theory, Stuttgart, Germany
\thanks{An extended version of this paper is available in \url{https://arxiv.org/abs/2109.14715}  \cite{eskandar2021usis}. Code is available at \url{https://github.com/GeorgeEskandar/USIS-Unsupervised-Semantic-Image-Synthesis}}
}

\begin{document}

\maketitle
 
\begin{abstract}
Semantic Image Synthesis (SIS) is a subclass of image-to-image translation where a semantic layout is used to generate a photorealistic image. State-of-the-art conditional Generative Adversarial Networks (GANs) need a huge amount of paired data to accomplish this task while generic unpaired image-to-image translation frameworks underperform in comparison, because they color-code semantic layouts and learn correspondences in appearance instead of semantic content. Starting from the assumption that a high quality generated image should be segmented back to its semantic layout, we propose a new Unsupervised paradigm for SIS (USIS) that makes use of a self-supervised segmentation loss and whole image wavelet based discrimination. Furthermore, in order to match the high-frequency distribution of real images, a novel generator architecture in the wavelet domain is proposed. We test our methodology on 3 challenging datasets and demonstrate its ability to bridge the performance gap between paired and unpaired models. 
\end{abstract}

\vspace{-1.3em}
\section{Introduction}
\label{sec:intro}
\vspace{-0.7em}
Semantic image synthesis (SIS) is the task of generating high resolution images from user-specified semantic layouts. SIS opens the door to an extensive range of applications such as content creation and semantic manipulation by editing, adding, removing or changing the appearance of an object. By allowing concept artists and art directors to brainstorm their designs efficiently, it can play a pivotal role in graphics design. In addition, SIS can be used as a data augmentation tool for deep learning models, by generating training data conditioned on desired scenarios which might be hard to capture in real-life (e.g. corner cases in autonomous driving like accidents). One of the first SIS frameworks \cite{wang2018high} was an adaptation of GANs \cite{ Goodfellow2014GenerativeAN, Karras2018ProgressiveGO, Karras2019ASG, Karras2020AnalyzingAI, isola2017image, Gal2021SWAGANAS}. SIS with SPatially Adaptive DEnormalization or SPADE \cite{park2019semantic} was proposed as an enhanced supervised methodology tailored specifically to suit SIS. Since then, a series of progressively enhanced frameworks were introduced \cite{wang2018high, park2019semantic, tan2020rethinking, liu2019learning, Zhu2020SEANIS}. Most notably, OASIS \cite{schonfeld2021you} extends the SPADE framework with a novel discriminator design and achieves state-of-the-art in supervised SIS. 

However, the problem of semantic image synthesis has mostly been addressed in a supervised setup. Although state-of-the-art methods \cite{wang2018high, schonfeld2021you, park2019semantic, tan2020rethinking, liu2019learning, Zhu2020SEANIS} can produce visually appealing high resolution images, they still depend on the availability of annotated training data which is expensive to acquire. For instance, the average annotation time for one image in the Cityscapes dataset is 1 hour \cite{cordts2016cityscapes}. 

Unpaired conditional GAN frameworks \cite{zhu2017unpaired, huang2018multimodal, lee2018diverse, park2020cut, fu2019geometry, benaim2017one, taigman2017unsupervised, shrivastava2017learning, bousmalis2017unsupervised, amodio2019travelgan, zhang2019harmonic} can be used to bypass the need for paired training data in SIS, but they suffer from several drawbacks: (1) instead of using one-hot encoding, these models color-code each class in the input semantic layout, which creates an artificial mapping between layouts and images, (2) the unsupervised losses in these works force relationships between the labels and images that do not preserve the semantic content in the case of SIS and (3) the normalization layers in the architecture of unsupervised models wash away the semantic labels as noted in \cite{park2019semantic}. Consequently, the generated samples from these models suffer from poor quality, especially when the number of classes in the training dataset is too big.

In this work, we propose an unsupervised framework which can synthesize realistic images from labels without the use of paired data. This is a first step towards bridging the performance gap between paired and unpaired frameworks. By virtue of its design, the unpaired setting can help eliminate dataset biases and push the model towards a better multimodal generation. The proposed Unsupervised SIS (USIS) \cite{eskandar2021usis} is a paradigm which involves an adversarial training between a generator and a whole image wavelet-based discriminator, and a cooperative training between the generator and a UNet segmentation network \cite{schonfeld2020u}. More precisely, the discriminator fosters the generator to match the distribution of the real images while the UNet gives a pixel-level feedback to the generator by means of a cross entropy loss. In addition, upon observing that convolutional networks are biased towards low-frequencies \cite{Chen2021SSDGANMT, Durall2020WatchYU, Dzanic2020FourierSD, Gao2016AHW, Liu2019MultiLevelWC, Williams2018WaveletPF, Liu2020WaveletBasedDN, Kang2017ADC, Liu2019AttributeAwareFA, Wang2020MultilevelWG, Huang2019WaveletDG}, we provide the wavelet decomposition of the real and fake images as input to our discriminator and we propose a novel wavelet-based generator architecture in order to approximate the high-frequency distribution of real images. We perform extensive experiments on 3 image datasets (COCO-stuff \cite{caesar2018coco}, Cityscapes \cite{cordts2016cityscapes} and ADE20K \cite{zhou2017scene}) in an unpaired setting to showcase the ability of our model to generate a high diversity of photorealistic images.
\section{Method}
\label{sec:method}
\begin{figure}
    \centering
    \begin{minipage}{\linewidth}
    \includegraphics[width=\textwidth]{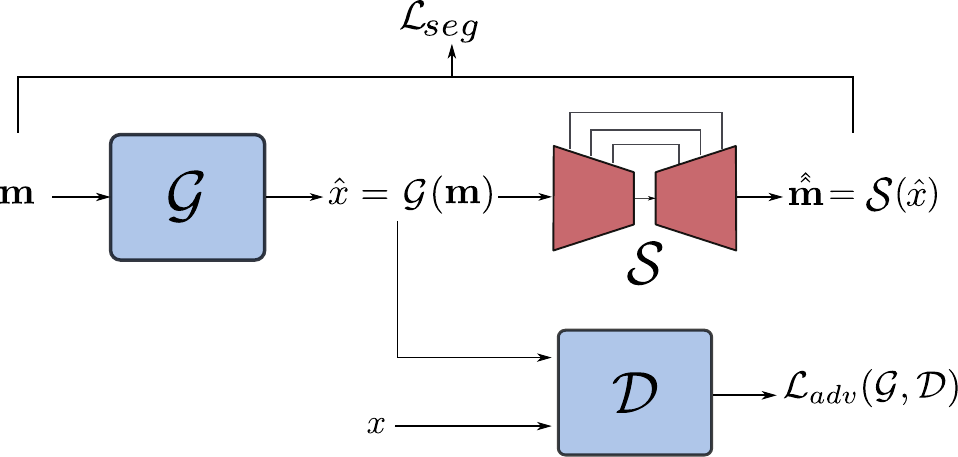}
    \end{minipage}
    \caption{The proposed unsupervised SIS paradigm}
    \label{fig:semantic_consistency}
\end{figure}

In SIS, we seek to synthesize an RGB image $\textbf{x}$ from a semantic mask $\textbf{m}$ (one-hot encoded) with $C$ classes in an unsupervised way. To achieve this, we propose the USIS framework which consists of three parts: (1) a waveletSPADE Generator $\mathcal{G}$, (2) a wavelet Discriminator $\mathcal{D}$ and (3) a UNet segmentation network $\mathcal{S}$. The generator generates a synthesized image $\hat{x}$ from the semantic map $\mathbf{m}$, the discriminator makes a decision whether the generated image is real or fake while the segmentation network tries to segment the generated image back to the mask $\mathbf{m}$. $\mathcal{S}$ only observes the generated images unlike the discriminator which sees real and generated images. In the following, we are going to introduce each of the 3 components in detail. The paradigm is depicted in Figure \ref{fig:semantic_consistency}.
\vspace{-2em}
\subsection{Semantic consistency loss}

A photorealistic image contains different objects which have different appearances and semantic meanings. Starting from this assumption, we impose a semantic consistency constraint to force the generator to produce images that lie on the real image manifold. A class-balanced segmentation loss $\mathcal{L}_{seg}$ between the synthesized images and the input labels is proposed to punish the inseparability between regions belonging to different semantic labels. This loss is self-supervised because it uses the input semantic layout as the groundtruth for segmentation, similar to autoencoders. The self-supervised loss pushes the generator to synthesize small classes and achieve a better semantic alignment. It can be expressed as: 
\begin{equation}
    \begin{aligned}
    \mathcal{L}_{seg} = - \mathbb{E}_{\textbf{m} } \left[ \sum_{c=1}^{C}\alpha_{c}\sum_{i,j}^{H \times W}\textbf{m}_{c,i,j} \log(\mathcal{S}(\mathcal{G}(\textbf{m}))_{c,i,j}) \right]    \end{aligned}
\end{equation}
The class-balancing weights $\alpha_{c}$ are proportional to the inverse of the per-pixel class-frequency.
\begin{ceqn}
    \begin{align}
    \alpha_{c} = \frac{H \times W}{ \sum_{i,j}^{H \times W} \mathbb{E}_{\textbf{m}} \left[ \mathbbm{1}[\textbf{m}_{c,i,j}] \right] } 
    \end{align}
\end{ceqn}
The overall loss can be expressed as $\mathcal{L}_{seg} + \lambda \mathcal{L}_{adv}$, where $\lambda$ is a hyperparameter.
\vspace{-1em}
\subsection{Whole image wavelet-based discrimination}
\begin{figure}[t]
    \centering
    \begin{minipage}{0.4\linewidth}
    \centering
    \includegraphics[width=0.6\textwidth]{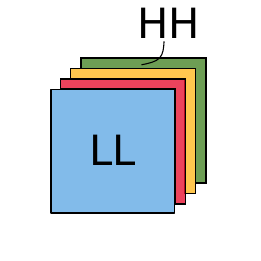}
    \subcaption{Channelwise-DWT}
    \label{fig:DWT:channel}
    \end{minipage}
    \qquad
    \begin{minipage}{0.4\linewidth}
    \centering
    \includegraphics[width=0.6\textwidth]{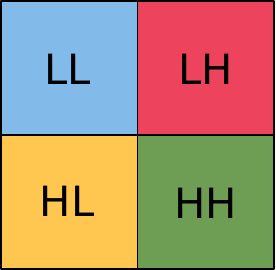}
    \subcaption{Spatial-DWT}
    \label{fig:DWT:spatial}
    \end{minipage}
    \caption{Different arrangement of the wavelet subbands inside the network}
    \vspace{-2em}
    \label{fig:DWT}
\end{figure}

The discriminator is an essential part of the framework because it is responsible for capturing the data statistics. Most importantly, it prevents the generator from learning trivial mappings (like identity mapping) that minimize the self-supervised segmentation loss. We propose to use whole image wavelet based discrimination for this task. More specifically, we incorporate the SWAGAN discriminator architecture, which was previously proposed in \cite{Gal2021SWAGANAS} to enhance the texture of generated images. By allowing the discriminator to process the Discrete Wavelet Transform (DWT) of the image, the higher frequencies are not entirely lost in the downsampling layers of the discriminator. Moreover, small objects in the spatial domain have wavelet coefficients with bigger magnitude, which are now observed by the discriminator and have a bigger contribution in $\mathcal{L}_{adv}$.
Formally, we define channelwise-DWT as the operation that takes a tensor of dimensions $(c \times w \times h)$ and outputs wavelet coefficients of dimensions $(4c \times w/2 \times h/2)$, where each input channel is transformed into 4 subbands of lower resolution, which are concatenated channelwise. Spatial-DWT refers to the concatenation of the subbands along the height and width of the features, yielding a tensor of the same dimension as the input. The Inverse Wavelet Transform (IWT) is defined as the inverse operation, which maps form the wavelet domain to the spatial domain. Unless otherwise stated, DWT and IWT refer to the channelwise-DWT and channelwise-IWT respectively. The 2 arrangements can be seen in Figure \ref{fig:DWT}. More details about the architecture of the utilized discriminator can be found in our prior publication \cite{eskandar2021usis}.

\vspace{-1em}
\subsection{Wavelet-based generator design}
In this subsection, we present a novel generator architecture: the waveletSPADE which is built upon the SPADE generator but designed to generate wavelet coefficients directly instead of pixels. More concretely, two modules are introduced: the wavelet upsample (WU) and the pixelSPADE (PS).
\begin{figure*}[t]
    \centering
    \begin{minipage}{\textwidth}
    \includegraphics[width=\textwidth]{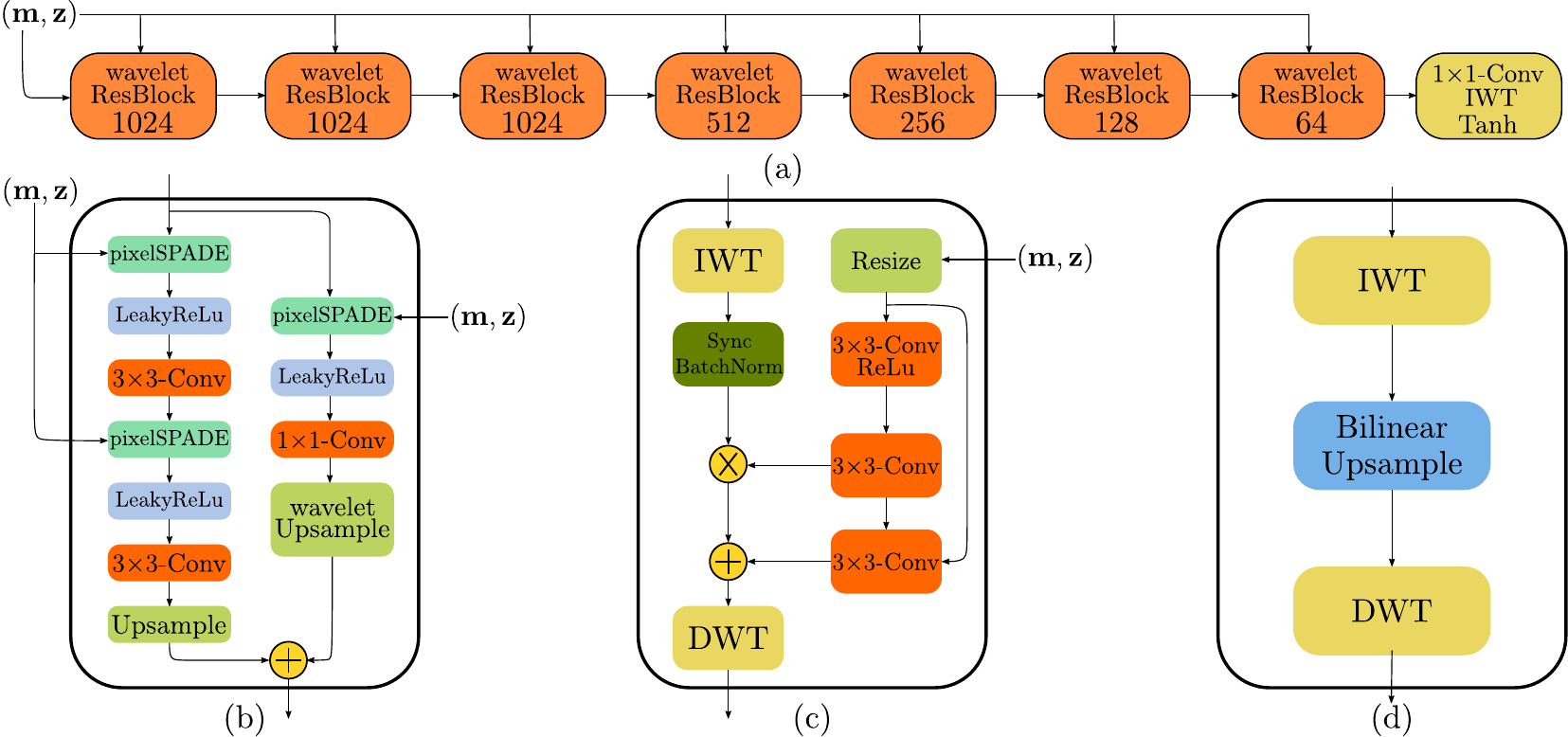}
    \end{minipage}
    \caption{The proposed waveletSPADE generator architecture (a) builds upon the OASIS generator \cite{schonfeld2021you} by replacing the ResBlocks with the depicted waveletResBlock (b) which encompasses the novel pixelSPADE (c) as well as the WU blocks (d). The number of channels per layer is shown in each waveletResBlock. }
    \label{fig:generator}
    \vspace{-1em}
\end{figure*}
The waveletSPADE architecture inherits its decoder-structure from SPADE but learns all the features in the wavelet domain. The input is the one-hot encoded segmentation map $\mathbf{m}$ concatenated with a 3D noise tensor, $\mathbf{z}$. Introduced in OASIS \cite{alharbi2020disentangled, schonfeld2021you}, 3D noise is a technique to learn multimodal generation by injecting structured noise in multiple layers of the network. The tensor is formed by sampling a latent vector of dimension $Z$ from a zero-mean unit-variance Gaussian distribution and propagating it at each pixel of the segmentation map resulting in a shape of ($(C+Z) \times H \times W$). The architecture and the blocks are shown in Figure \ref{fig:generator}.

{\bf Wavelet Upsampling: } In contrast to the original SPADE architecture, ResBlocks and upsampling layers are not interleaved. Instead, we move the upsampling layers inside the ResBlocks but use different scaling methods inside each branch: the identity branch is upsampled using a waveletUpsample (WU) block, while the residual branch is upsampled with a nearest-neighbour interpolation. The motivation for such a design choice is that we would like to preserve identity mappings in the shortcut connection of the ResBlock, as they are necessary to channel proper gradients to the early layers of the network which produce coarse features. The waveletUpsample layer, inspired by SWAGAN\cite{Gal2021SWAGANAS}, consists of an IWT, followed by a bilinear upsampling in the spatial domain and a DWT. This stands in contrast to applying transposed convolutions directly on high features inside the network because this would ignore that the 4 subbands of DWT contain different frequency information and would introduce low-frequency artifacts in the generated image. On the other hand, we use nearest-neighbour upsampling in the residual branch because it preserves the sharpness of the wavelet coefficients (otherwise high value coefficients could be attenuated by linear interpolation techniques).

The {\bf pixelSPADE} layer replaces the original SPADE layer in all the ResBlocks. Since the style of an image correlates to the mean and standard deviation of the features inside the network, the original SPADE was designed to generate demodulation parameters per pixel. However, applying SPADE to each subband of the DWT is suboptimal as SPADE was designed to be applied in the spatial domain, not on the frequency components of the image. Therefore, we propose to use pixelSPADE which consists of an IWT operation, a SPADE layer and a DWT. The block thus offers the advantage of preserving the frequency content while applying the style to the features in the spatial domain.
\vspace{-1.5em}
\section{Experiments}
\label{sec:experiments}
\vspace{-1em}
\begin{figure*}[t]
    \captionsetup[subfloat]{position=top, labelformat=empty, skip=0pt}

    \begin{minipage}{\textwidth}
    \subfloat[Label]{\includegraphics[width=  0.16\textwidth, height=  0.12\textwidth]{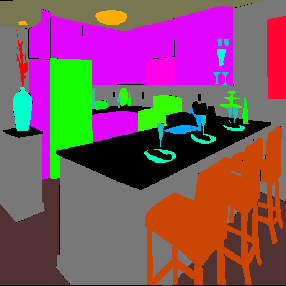}} \hfill
    \subfloat[CycleGAN \cite{zhu2017unpaired}]{\includegraphics[width=  0.16\textwidth, height=  0.12\textwidth]{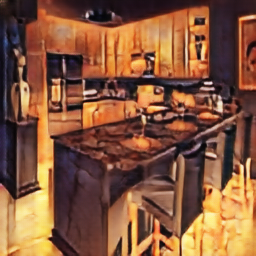}} \hfill
    \subfloat[CUT \cite{park2020cut}]{\includegraphics[width=  0.16\textwidth, height=  0.12\textwidth]{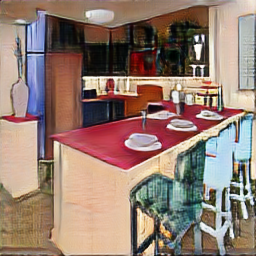}} \hfill
    \subfloat[USIS]{\includegraphics[width=  0.16\textwidth, height=  0.12\textwidth]{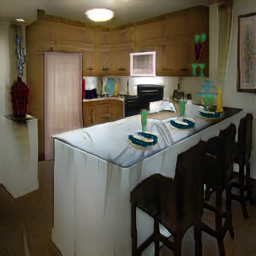}} \hfill
    \subfloat[OASIS \cite{schonfeld2021you}]{\includegraphics[width=  0.16\textwidth, height=  0.12\textwidth]{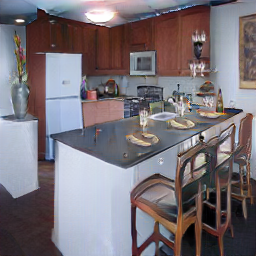}} \hfill
    \subfloat[Groundtruth]{\includegraphics[width=  0.16\textwidth, height=  0.12\textwidth]{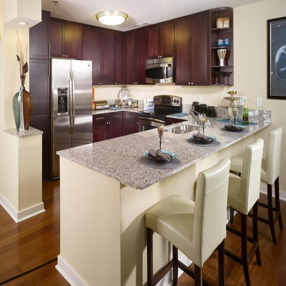}}
    \vspace{-1em}
    \subfloat[]{\includegraphics[width=  0.16\textwidth, height=  0.12\textwidth]{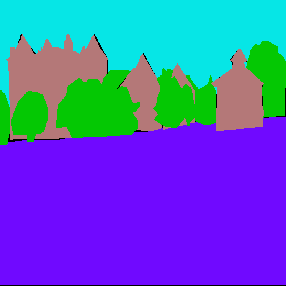}} \hfill
    \subfloat[]{\includegraphics[width=  0.16\textwidth, height=  0.12\textwidth]{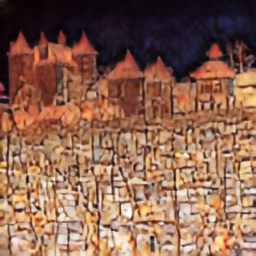}} \hfill
    \subfloat[]{\includegraphics[width=  0.16\textwidth, height=  0.12\textwidth]{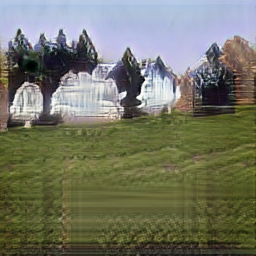}} \hfill
    \subfloat[]{\includegraphics[width=  0.16\textwidth, height=  0.12\textwidth]{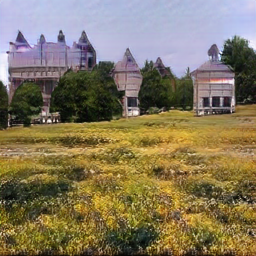}} \hfill
    \subfloat[]{\includegraphics[width=  0.16\textwidth, height=  0.12\textwidth]{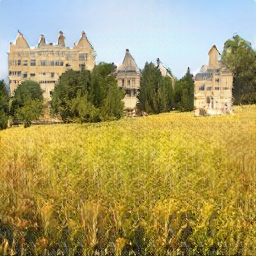}} \hfill
    \subfloat[]{\includegraphics[width=  0.16\textwidth, height=  0.12\textwidth]{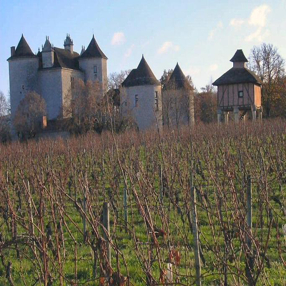}}
    \vfill
    \end{minipage}
   \caption{Qualitative comparison against state-of-the-art unpaired SIS on ADE20K. OASIS is the state-of-the-art in paired SIS.}
   \vspace{-1.5em}
   \label{fig:ade20k_paper_results}
\end{figure*}

We conduct We conduct our experiments for unsupervised SIS on 3 datasets: Cityscapes\cite{cordts2016cityscapes}, COCO-stuff\cite{caesar2018coco} and ADE20K\cite{zhou2017scene}. Cityscapes contains street scenes in German cities with pixel-level annotations of 19 classes. It is widely used for vision tasks in autonomous driving and contains 3000 training images and 500 test images. ADE20K and COCO-stuff are more challenging datasets because they offer a high diversity of indoor and outdoor scenes; and they have a lot of semantic classes. COCO-stuff has 182 classes while ADE20K has 150 classes. These 3 datasets are the standard benchmark in supervised SIS. We use a resolution of 256 $\times$ 512 for Cityscapes and  a resolution of 256 $\times$ 256 for ADE20K and Cocostuff. In previous works on unpaired GANs, the semantic image synthesis experiments were only performed on Cityscapes and/or datasets with small number of classes. The standard evaluation metrics for this task are utilized to measure both the quality and diversity of generated images. Specifically, we use the Frechet Inception Distance or FID\cite{heuselttur2017}, to assess both quality and diversity. We also follow SPADE \cite{park2019semantic} and run pretrained semantic segmentation models \cite{yu2017dilated, chen2014semantic, xiao2018unified} on the generated images and report the mean Intersection-over-Union (mIoU) to evaluate the semantic alignment and visual quality. We also perform an ablation study on Cityscapes to analyze the effect of the different components in the generator architecture. All the experiments are conducted using the OASIS generator (SPADE generator + 3D noise), the wavelet-based discriminator and the UNet. The batchsize is 8. Adding an IWT after the generator is referred to as OASIS + IWT. WU$^{*}$ refers to the proposed wavelet upsample in \cite{Tsai2006ImageUU}. 
\begin{table}

	 %

	\setlength{\tabcolsep}{0.2em}
	\renewcommand{\arraystretch}{0.95}
	\centering
	
	\begin{tabular}{l|cc|cc|cc}
    	\multirow{2}{*}{\normalsize{} Method } & \multicolumn{2}{c|}{\normalsize{} Cityscapes} &  \multicolumn{2}{c|}{\normalsize{} ADE20K} & \multicolumn{2}{c}{\normalsize{} COCO-stuff} 
    	\tabularnewline
    	  &  \normalsize{} FID$\downarrow$  & \normalsize{} mIoU$\uparrow$ &  \normalsize{} FID$\downarrow$  & \normalsize{} mIoU$\uparrow$ &  \normalsize{} FID$\downarrow$  & \normalsize{} mIoU$\uparrow$ \tabularnewline
    	
    	\hline 
    	
    	{\small{CycleGAN} \cite{zhu2017unpaired} } &   \normalsize{87.2} & \normalsize{24.5}  &  \normalsize{96.3} & \normalsize{5.4} &  \normalsize{104.7} & \normalsize{2.08} \tabularnewline

    	{\small{MUNIT} \cite{huang2018multimodal}  } &   \normalsize{84} & \normalsize{8.2} &   \normalsize{n/a} & \normalsize{n/a}  &  \normalsize{n/a} & \normalsize{n/a} \tabularnewline

    	{\small{DRIT} \cite{lee2018diverse}  } &  \normalsize{164} & \normalsize{9.5} &  \normalsize{132.2} & \normalsize{0.016} &  \normalsize{135.5} & \normalsize{0.008} \tabularnewline

    	 {\small{Distance} \cite{benaim2017one}  } &   \normalsize{78} & \normalsize{17.6} &  \normalsize{80} & \normalsize{0.035} &  \normalsize{92.4} & \normalsize{0.014 } \ \tabularnewline

    	{\small{GCGAN} \cite{fu2019geometry} } &   \normalsize{80} & \normalsize{8.4} &   \normalsize{92} & \normalsize{0.07} &  \normalsize{99.8} & \normalsize{0.019} \tabularnewline

    	\small{CUT} \cite{park2020cut} &   
    	\normalsize{57.3} & \normalsize{29.8} &  \normalsize{79.1}  &\normalsize{6.9} &  \normalsize{85.6} & \normalsize{2.21}  \tabularnewline
    	
    	 \hline
    	
    	\small{} USIS	&   \normalsize{\textbf{50.14}} & \normalsize{\textbf{42.32}} & \normalsize{\textbf{34.5}} & \normalsize{\textbf{16.95}} & \normalsize{\textbf{28.6}} & \normalsize{\textbf{13.4}} \tabularnewline
		
		\end{tabular}
	\caption{Comparison against state-of-the-art unpaired GANs.}
	\label{table:comp_with_sota}
	\vspace{-1.5em}
\end{table}

\vspace{-1.7em}
\section{Results}
\label{sec:results}
\vspace{-1em}
The quantitative comparison is reported in Table \ref{table:comp_with_sota} and some samples of the generated images are shown in Figure \ref{fig:ade20k_paper_results}. CycleGAN results in the most unrealistic images, as it is unable to math the color and texture distribution of the real images although it is able to produce objects with clear boundaries. This is reflected in the suboptimal FID scores but relatively high mIoU scores in all 3 datasets in Table \ref{table:comp_with_sota}. More recent frameworks like DistanceGAN or GCGAN have low FID scores but the objects might often be indistinguishable. CUT has the best FID and mIoU combination among the baselines, but suffers from a deterioration when the number of classes is high. The proposed USIS model is able to bypass this problem, and generates photorealistic images on the challenging datasets by virtue of the semantic consistency loss and its wavelet-based discriminator and generator designs.    
\begin{table}[h]
    \begin{tabular}{l|cc}
         Method & FID & mIoU  \\ 
         \hline
         OASIS                         & 52.19 & \textbf{42.8}  \\
         OASIS + IWT                   & 50.52 & 40.27 \\
         OASIS + IWT + WU              & 51.37 & 39.88 \\
         OASIS + IWT + WU$^{*}$        & 54.78 & 35.74 \\
         OASIS + Spatial-IWT + WU      & 55.44 & 28.22 \\
         OASIS + Spatial-IWT + WU + PS & 52.69 & 40.66 \\
         OASIS + IWT + WU + PS         & \textbf{50.14} & 42.32 \\
         
    \end{tabular}
\caption{Ablation study on Cityscapes.}
\label{tab:ablation}
\vspace{-1.5em}
\end{table}

The results of the ablation study are reported in Table \ref{tab:ablation}. The simple addition of a channelwise IWT at the end of the OASIS generator enhances the FID as the generator is better able to generate more high-frequency content and produce better texture. However, this comes at the expense of the semantic alignment. Replacing upsampling layers by WU layers alone seems to slightly worsen the results, leading to more misalignment. It's the introduction of the PixelSPADE layer that restores the good alignment results with a negligible degradation while lowering the FID, thus approaching the state-of-the-art supervised performance (47.7 in OASIS \cite{schonfeld2021you}). We have also experimented with Spatial-DWT and IWT leading to suboptimal results. In contrast, representing frequency information in the channels helps the network learn more useful features.    
\vspace{-1.5em}
\section{Conclusion}
\vspace{-1em}
\label{sec:conclusion}
In this work, a framework for semantic image synthesis in an unpaired setting was proposed (USIS). It deploys a waveletSPADE generator along with a UNet and an unconditional whole image wavelet-based discriminator. The UNet fosters class separability and content preservation while the discriminator matches the color and texture distribution of real images. The effectiveness of the proposed framework in the semantic image synthesis was shown on 3 challenging datasets: Cityscapes, ADE20K and Cocostuff. USIS outperformed prior unpaired GANs while approaching the performance of supervised frameworks. An ablation study was performed to analyze the role of the different components in the unsupervised paradigm. 
\vspace{-1em}
\section*{\centering \normalsize Acknowledgement}
\label{sec:ack}
\vspace{-1em}
\setstretch{0.6}
{\small The research leading to these results is funded by the German Federal Ministry for Economic Affairs and Energy within the project "AI Delta Learning".  The authors would like to thank the consortium for the successful cooperation.}

\clearpage

\begingroup
\setstretch{0.6}
\setlength\bibitemsep{0pt}
\printbibliography
\endgroup
\end{document}